# Wasserstein Distance-Weighted Adversarial Network for Cross-Domain Credit Risk Assessment

Mohan Jiang
New York University
New York, USA
mj2589@nyu.edu

Jiating Lin
Brown University
Providence, USA
linjiating96@gmail.com

Hongju Ouyang
Columbia University
New York , USA
hongjuoy0612@gmail.com

Jingming Pan
University of Pennsylvania
Philadelphia, USA
panjing@alumni.upenn.edu

Siyuan Han
Columbia University
New York, USA
siyuan.han@columbia.edu

Bingyao Liu*
University of California, Irvine
Irvine, USA
*Corresponding author:
bingyl5@uci.edu

*Abstract*—This paper delves into the application of adversarial domain adaptation (ADA) for enhancing credit risk assessment in financial institutions. It addresses two critical challenges: the cold start problem, where historical lending data is scarce, and the data imbalance issue, where high-risk transactions are underrepresented. The paper introduces an improved ADA framework, the Wasserstein Distance Weighted Adversarial Domain Adaptation Network (WD-WADA), which leverages the Wasserstein distance to align source and target domains effectively. The proposed method includes an innovative weighted strategy to tackle data imbalance, adjusting for both the class distribution and the difficulty level of predictions. The paper demonstrates that WD-WADA not only mitigates the cold start problem but also provides a more accurate measure of domain differences, leading to improved cross-domain credit risk assessment. Extensive experiments on real-world credit datasets validate the model's effectiveness, showcasing superior performance in cross-domain learning, classification accuracy, and model stability compared to traditional methods.

## I. Introduction

Credit risk assessment is an essential process for banks and financial institutions to manage credit risk. In recent years, methods represented by deep learning have been applied to credit risk prediction [1]. Recent studies have proposed solutions based on transfer learning to predict credit risk, yet there are still some issues that urgently need improvement. Firstly, new business development faces the dilemma of a cold start at the beginning, where credit risk assessment often lacks or has limited historical lending data. How to use transfer learning for cross-domain risk identification is a challenge. Secondly, credit risk data is often imbalanced, with high-risk transaction behavior accounting for only a very small part of all transaction behaviors, which brings difficulties and challenges to the establishment of risk identification models [2].

This paper takes adversarial domain adaptation as the research objective and explores customer default prediction in the field of credit risk. Firstly, it introduces the methods, principles, and bottlenecks of adversarial domain adaptation, as well as the data imbalance issues present in the field of credit risk. Secondly, in response to the aforementioned challenges, the paper optimizes the traditional adversarial domain adaptation methods in two aspects: the adversarial domain adaptation algorithm based on Wasserstein distance and the weighted label classification loss function [3]. The former introduces the distribution distance metric—Wasserstein distance, using the Wasserstein distance estimation to minimize the distribution differences between the source domain and the target domain. The former introduces the distribution distance metric—Wasserstein distance—using its estimation to minimize the distribution differences between the source domain and the target domain. This method has been successfully applied in areas like computer vision to enhance model performance across varied datasets [4-7]. The latter proposes a dynamically balanced weighted loss function to regulate the imbalance between minority and majority class samples during the training process. By proposing an improved adversarial domain adaptation network (WD-WADA) to address the cold start problem and the imbalance of risky user data in the field of credit risk. Finally, the effectiveness of the constructed model is verified on real credit datasets.

The article introduces an enhanced adversarial domain adaptation network utilizing Wasserstein distance to improve cross-domain alignment and mitigate issues like gradient disappearance in adversarial learning. A novel weighting strategy is also proposed to address data imbalance, modifying traditional weighting methods to stabilize model training. The development of the WD-WADA model specifically targets credit risk assessment, effectively managing cold start problems and aligning closely with real-world financial market dynamics. Extensive evaluations demonstrate that WD-WADA surpasses conventional models in terms of classification accuracy and stability, highlighting its effectiveness in practical applications.

## II. RELATED WORK

The application of adversarial domain adaptation (ADA) in addressing challenges such as the cold start problem and data imbalance in credit risk assessment builds upon various deep learning methodologies. These approaches serve as the foundation for the development of the Wasserstein Distance-Weighted Adversarial Domain Adaptation Network (WD-WADA), which enhances credit risk prediction by effectively aligning domain distributions and mitigating data imbalance.

One critical aspect of cross-domain learning is the optimization of deep learning models to improve performance across diverse financial datasets. Previous research has demonstrated the importance of extracting meaningful features and optimizing models for better predictive power, particularly in complex financial data environments [8]. These insights are relevant to WD-WADA, which seeks to enhance feature extraction and domain alignment to achieve accurate cross-domain credit risk assessment.

Addressing the detection of anomalies and irregular patterns in financial data has also been a key focus, especially when these patterns are underrepresented, similar to the imbalance of high-risk transactions in credit datasets. Techniques for anomaly detection provide useful methodologies for tackling the challenges posed by imbalanced data in ADA models [9]. WD-WADA builds on these approaches by incorporating an innovative weighted strategy that adjusts the model's focus on underrepresented classes based on their prediction difficulty and rarity.

Furthermore, advancements in adaptive methods for spatio-temporal data aggregation have contributed significantly to improving risk detection models [10]. These methods dynamically adjust to evolving patterns, similar to how WD-WADA uses Wasserstein distance to ensure domain alignment as data distributions change over time. This dynamic adaptation is essential in financial risk environments where behaviors can vary significantly between domains.

Additional work in improving the stability and training efficiency of neural networks through advanced numerical methods has been instrumental in refining deep learning architectures. The incorporation of higher-order numerical techniques to stabilize training processes mirrors the enhancements made in WD-WADA, particularly in addressing challenges like gradient vanishing during adversarial learning [11]. Moreover, generating realistic data from limited samples has been explored as a solution to the cold start problem, where historical data is scarce. Techniques such as adversarial learning for generating synthetic data have been applied in various domains [12] and are directly relevant to the credit risk context, where WD-WADA leverages adversarial domain adaptation to align sparse data distributions across different financial markets.

Optimized training strategies, including improvements in gradient descent methods, also play a pivotal role in enhancing model robustness and convergence. These techniques ensure that adversarial networks like WD-WADA maintain stability during the domain adaptation process, particularly when handling imbalanced datasets and complex domain shifts [13]. In addition, cross-domain predictive models that operate effectively across heterogeneous datasets provide valuable insights into how knowledge can be transferred between domains, a central concept in ADA [14]. The WD-WADA framework incorporates similar principles to align credit risk data from diverse sources, ensuring that domain differences are minimized, and predictions are accurate.

Finally, techniques for embedding and aligning complex relationships between data modalities also inform the development of WD-WADA [15]. Effective alignment of features across different domains is crucial for improving cross-domain credit risk assessment, and methodologies focusing on embedding strategies contribute to the precision of this alignment.

## III. METHOD

### A. Feature Mapping Based on CNN

In credit scenarios, almost all data consists of tabular statistical features. We collect user data while adhering to security and privacy policies. We utilize a feature extractor to process these raw inputs as shown in Figure 1. The encoded features from the source domain and target domain are represented with the same dimensional output as follows:

$$z_i = G_f(x_i), i = 1 \ldots n, x_i \in X^s, X^t$$

First we train the domain data using CNN. A CNN model is pre-trained on the labeled dataset from the source domain $X^s$:

The convolutional layer includes a filter $w \in \mathbb{R}^k$ and a bias $b \in \mathbb{R}$ to compute a new feature. The output feature $v_i$ is obtained through the filter w and a nonlinear activation function C, expressed as:

$$v_i = \mathrm{T}(w u_j + b)$$

To compute the difference between the predicted label $\tilde{y}_i^s$ is and the true label from the source domain $y_i^s$, a cross-entropy function is used to calculate the loss:

$$L_y = \min{}_{G_f G_y} \mathbb{E}_{(x',y') \sim (x',y')} \left[ -\sum_{c=1}^{c} \mathbb{I}_{[c=y']} \log G_y(G_f(X^s)) \right]$$

### B. Domain Adaptation Based on Wasserstein Distance

The next challenge is to address the distribution differences between the source dataset and the target dataset. To tackle this issue, we utilize Wasserstein distance [16], learning invariant feature representations in a common latent space between two different feature distributions through adversarial training. In this paper, the domain discriminator is designed to align the marginal distributions between the source and target domains. The overall idea of the domain discriminator follows the Domain-Adversarial Neural Network (DANN) approach, which can be trained to estimate

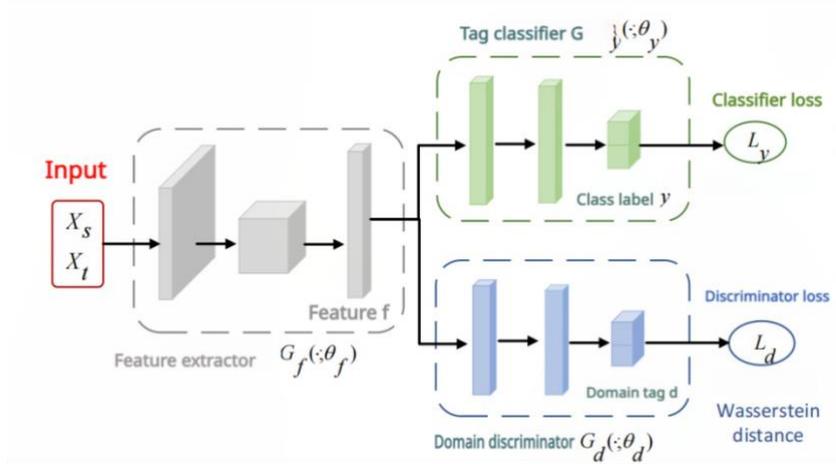

Figure 1. Flow chart of weight mask principle

Wasserstein distance as follows:
$$L_{vd} = \max_{G_d} \mathbb{E}_{x'\sim p^t}\left[G_d\left(G_f(x^t)\right)\right] - \mathbb{E}_{x^s\sim p^t}\left[G_d\left(G_f(x^s)\right)\right]$$

The method proposed using weight clipping to satisfy the Lipschitz constraint. This straightforward approach clips the parameters of each layer of the discriminator's neural network to a fixed interval, which may result in some parameters being set to 1 or -1, inevitably limiting the discriminator's fitting ability. Gulrajani et al. pointed out that weight clipping in traditional WGANs [17] can restrict model stability and cause gradient vanishing or explosion in adversarial training, and that the weight clipping operation is cumbersome. Therefore, it is recommended to satisfy the Lipschitz continuity constraint through gradient penalty. To prevent the aforementioned issues, this paper constructs a gradient penalty term added to the discriminator's loss function to impose a softer penalty on the discriminator's parameters.

Assuming the intermediate layer feature representation of the discriminator that needs to be penalized is h, the gradient can be computed to obtain the weights of that layer's discriminator, fixing the value to 1, as shown in the following equation:
$$L_{\text{grad}} = \mathbb{E}\left(\|\nabla_h D(h)\|_2 - 1\right)^2,$$
where $\nabla_h D(h)$ represents the gradient of h with respect to the discriminator output D(h). Adding this to the objective function allows for the limitation of the discriminator's weights, thus automatically satisfying the Lipschitz constraint. The empirical Wasserstein distance estimation formula is as follows:
$$\max_{G_d}\{L_{\text{vd}} - \rho L_{gmad}\}$$
$$= \max_{G_d}\left\{\mathbb{E}_{x'-p^p}\left[G_d\left(G_f(x^t)\right)\right] - \mathbb{E}_{x^s-p^p}\left[G_d\left(G_f(x^s)\right)\right] - \rho L_{\text{gud}}\right\},$$

The adversarial objective based on Wasserstein distance is defined as:
$$L_d = \min_{G_f}\max_{G_d}\{L_{wd} - \rho L_{g\text{ rad}}\}.$$

### C. Weighted Strategy for Label Classifier

The label classifier $G_y$ is trained to identify the labels of input samples in the source domain. Consequently, the supervisory information from $\mathcal{D}_s$ can be utilized by $\mathcal{D}_t$. Since the training objective is a binary classification task, where sample labels $y \in \{0,1\}$, minority class samples are labeled as 1 (positive samples), and majority class samples are labeled as 0 (negative samples). The training objective function is the cross-entropy loss, which can be expressed as:
$$L_y = -\frac{1}{n_s}\sum_{x_i\in\mathcal{D}_s}[y_i\cdot\log p_i + (1-y_i)\cdot\log(1-p_i)]$$
For a single sample, the cross-entropy loss function is:
$$-\log p_i, y_i = 1$$
$$-\log(1-p_i), y_i = 0$$
where $p_i = G_y(G_f(\mathbf{x}_i))$ is the predicted probability output by the label classifier.

The label classification network applies the same weight to all instances. However, since the minority class instances are our core target for prediction, and given that the features of the minority class significantly impact classification, we propose a sample weighting strategy. This strategy adjusts the importance of samples from two aspects: the proportion of majority to minority class samples in the overall sample set and the difficulty of classifying the samples. The former allows for different weights on the losses of different classes, increasing the weight of minority class samples, as shown in Equation.

The latter is based on the concept of Focal Loss, which addresses the classification difficulty caused by sample imbalance. By reducing the loss of easily classified samples, the loss function focuses on hard-to-classify samples, which not only helps alleviate the sample imbalance issue but also improves the overall performance of the model.

The concepts of hard-to-classify and easy-to-classify samples are dynamic and change throughout the training process. The original Focal Loss is expressed as:

$$\text{FocalLoss} = -\frac{1}{n_s} \sum_{x_i \in D_s} [y_i(1-p_i)^r \log p_i + (1-y_i)p_i^r \log(1-p_i)]$$

This means that by lowering the weight of easy-to-classify samples, we indirectly increase the importance of hard-to-classify samples in the loss function, making it more inclined to train on those difficult samples.

## IV. EXPERIMENT

### A. Dataset

Lending Club is an intermediary service platform that provides credit loans, primarily focused on personal consumption loans and small business loans. Its credit lending rating dataset, LC, includes information such as the applicant's age, gender, marital status, education, loan amount, and applicant's financial situation. This information can be used to predict whether a loan application will default, thereby determining whether to grant the loan.

This study selects the LC dataset from 2007 to 2020, focusing specifically on credit card loan data. The credit card loan dataset contains 319,000 detailed records and 142 feature variables, with two target variables: default (charged off) and repayment (fully paid).

### B. Network Parameter

Feature Extractor Architecture Comprises two convolutional layers, two max-pooling layers, and two fully connected layers (FC1-FC2). The activation function used in the convolutional layers is ReLU. The feature extractor takes 38-dimensional structured data as input and outputs a 32-dimensional vector after undergoing multi-layer convolution processing. The parameter settings for each layer can be referred to in Table 1.

Table 1. Network structure of feature extractor

| Layer Name | Filters | Kernel Size | Stride |
|---|---|---|---|
| Convolutional Layer | 16 | 1x6 | 2 |
| Pooling Layer | - | 1x2 | 2 |
| Convolutional Layer | 32 | 1x6 | 2 |
| Pooling Layer | - | 1x2 | 2 |

### C. Experiment Results Analysis

This study uses three regions with sufficient data (CA, NY, TX) as the source domain dataset and a region with less data (UT) as the target domain data. A random sample of 20,000 records is drawn from the source domain, while 2,000 samples are randomly selected from the target domain, with 80% used for training and 20% for testing. Four models are utilized to predict the test set of the target domain, and the performance of WD-WADA, WD-ADA, and other traditional methods is compared using three evaluation metrics: Precision, F1-score, and AUC. The prediction results of the four models across three different transfer tasks are shown in Table 2.

Table 2. Performance of transfer tasks

| | Model / Metric | CNN | DANN | WD-ADA | WD-WADA |
|---|---|---|---|---|---|
| CA→UT | Precision | 0.65 | 0.77 | 0.81 | 0.85 |
| | F1-score | 0.56 | 0.72 | 0.79 | 0.78 |
| | AUC | 0.58 | 0.67 | 0.791 | 0.79 |
| NY→UT | Precision | 0.66 | 0.79 | 0.82 | 0.87 |
| | F1-score | 0.68 | 0.78 | 0.78 | 0.80 |
| | AUC | 0.67 | 0.77 | 0.76 | 0.78 |
| TX→UT | Precision | 0.61 | 0.83 | 0.83 | 0.84 |
| | F1-score | 0.57 | 0.79 | 0.80 | 0.82 |
| | AUC | 0.59 | 0.73 | 0.77 | 0.80 |

From Table 2, the WD-WADA model excels in two of the three transfer directions (NY→UT and TX→UT) across all metrics, and performs well in one metric for CA→UT. Compared to the CNN model, which uses only target domain samples, WD-WADA demonstrates substantial advantages in unsupervised classification, notably outperforming the CNN with no transfer learning by achieving a 28% improvement in precision across regions. Moreover, WD-WADA generally surpasses most models in transfer accuracy, with an average increase of 5%, except in the NY to UT direction where it slightly lags by less than 1%.

In individual metrics, WD-WADA shows superior performance: it achieves a Precision of 0.83, a 27.7% increase from the best previous benchmark of 0.65. Its maximum AUC is 0.80, up by 19.4% from the highest prior benchmark of 0.67. Similarly, the F1 score peaks at 0.82, an 18% improvement over the previous best of 0.68. These results confirm WD-WADA's effectiveness in utilizing weighting strategies and Wasserstein distance-based adversarial domain adaptation for feature transfer. Despite lacking domain adaptation, the CNN still delivers reasonable performance in credit risk feature transfer tasks due to its strong feature detection capability.

### D. Robustness Analysis

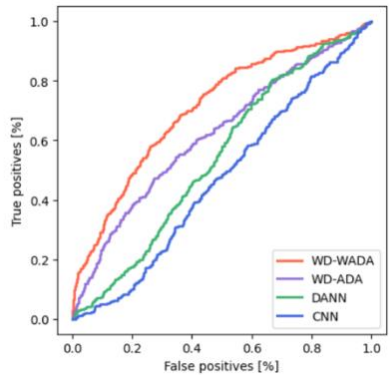

Figure 2. ROC Curve of Model

This paper also investigates the robustness of the proposed algorithm WD-WADA and compares it with CNN, DANN, and WD-ADA methods. By examining the range of changes in the accuracy of the transfer task, the robustness of the WA-WADA model is tested. The task of transferring from CA to UT was selected for this transfer task, and we ran the task five times, storing the AUC for each run. Figure 2. present the changes in the accuracy of the transfer task. The blue shading represents the 95% confidence interval of the AUC; the narrower the interval, the more stable the model during the experimental training process. It can be observed that the AUC of the adversarial domain adaptation method based on the Wasserstein distance is not only higher than the other two methods but also has a narrower 95% confidence interval than the other two methods. This confirms our motivation for using Wasserstein distance-based domain adaptation, which not only enhances the transfer capability of domain adaptation but also improves the instability of traditional domain adaptation network training. By employing our proposed algorithm, both the accuracy of feature transfer and the robustness of the model have been strengthened.

## V. CONCLUSION

This paper introduces a groundbreaking framework, the Wasserstein Distance-Weighted Adversarial Domain Adaptation Network (WD-WADA), which offers a novel approach to improving cross-domain credit risk assessment. The proposed model effectively addresses two key challenges: the cold start problem, where historical lending data is scarce, and the prevalent issue of data imbalance, particularly the underrepresentation of high-risk transactions. By incorporating the Wasserstein distance to measure and align distribution differences between the source and target domains, the model enhances cross-domain learning, mitigating the limitations of traditional adversarial domain adaptation approaches. Furthermore, WD-WADA introduces an innovative weighted strategy that adjusts class distribution based on sample difficulty and rarity, thus addressing the imbalances in high-risk credit data. This dual-focus approach not only improves prediction accuracy but also enhances model stability and performance across various scenarios. The experimental results on real-world credit datasets demonstrate that WD-WADA significantly outperforms conventional models in terms of classification accuracy, domain alignment, and robustness. This research not only provides a scalable and effective solution to pressing issues in credit risk assessment but also opens new avenues for applying adversarial domain adaptation in financial risk management. As such, it contributes valuable insights to the field and lays the groundwork for future exploration into AI-driven credit risk analysis.